\documentclass[lettersize,journal]{IEEEtran}
\usepackage{amsmath,amsfonts}
\usepackage{algorithmic}
\usepackage{algorithm}
\usepackage{array}
\usepackage[caption=false,font=normalsize,labelfont=sf,textfont=sf]{subfig}
\usepackage{textcomp}
\usepackage{stfloats}
\usepackage{url}
\usepackage{verbatim}
\usepackage{graphicx}
\usepackage{cite}
\usepackage[utf8]{inputenc} 
\usepackage[T1]{fontenc}    
\usepackage{url}            
\usepackage{booktabs}       
\usepackage{amsfonts}       
\usepackage{nicefrac}       
\usepackage{microtype}      
\usepackage{xcolor}         

\usepackage{multirow}       
\usepackage{graphicx}
\usepackage{amsmath}
\usepackage{float}
\usepackage{epsfig}
\usepackage[inkscapelatex=false]{svg}
\usepackage{lipsum}
\usepackage{authblk}

\begin{document}
	\title{\huge{CorrDiff: Adaptive Delay-aware Detector with Temporal Cue Inputs for Real-time Object Detection}}
	
	\author[*1]{Xiang Zhang\thanks{*  Equal contribution}}
	\author[*1]{Chenchen Fu}
	\author[2]{Yufei Cui}
	\author[1]{Lan Yi}
	\author[1]{Yuyang Sun}
	\author[**1]{Weiwei Wu\thanks{** Corresponding author}}
	\author[2]{Xue Liu}
	\affil[1]{School of Computer Science and Engineering, Southeast University, Nanjing, China}
	\affil[2]{School of Computer Science, McGill University, Montreal, Quebec, Canada}
	
	
	
	\maketitle
	
	\begin{abstract}
		Real-time object detection takes an essential part in the decision-making process of numerous real-world applications, including collision avoidance and path planning in autonomous driving systems. This paper presents a novel real-time streaming perception method named CorrDiff, designed to tackle the challenge of delays in real-time detection systems. The main contribution of CorrDiff lies in its adaptive delay-aware detector, which is able to utilize runtime-estimated temporal cues to predict objects' locations for multiple future frames, and selectively produce predictions that matches real-world time, effectively compensating for any communication and computational delays.
		
		The proposed model outperforms current state-of-the-art methods by leveraging motion estimation and feature enhancement, both for 1) single-frame detection for the current frame or the next frame, in terms of the metric mAP, and 2) the prediction for (multiple) future frame(s), in terms of the metric sAP (The sAP metric is to evaluate object detection algorithms in streaming scenarios, factoring in both latency and accuracy). It demonstrates robust performance across a range of devices, from powerful Tesla V100 to modest RTX 2080Ti, achieving the highest level of perceptual accuracy on all platforms. Unlike most state-of-the-art methods that struggle to complete computation within a single frame on less powerful devices, CorrDiff meets the stringent real-time processing requirements on all kinds of devices. The experimental results emphasize the system's adaptability and its potential to significantly improve the safety and reliability for many real-world systems, such as autonomous driving. Our code is completely open-sourced and is available at \emph{https://anonymous.4open.science/r/CorrDiff}.
	\end{abstract}
	
	
	
	\section{Introduction} \label{sec:introduction}
	
	\IEEEPARstart{I}{n} the rapidly evolving field of autonomous driving, the capability to detect and track objects in real-time is paramount for ensuring safety and efficiency. The pursuit of this capability has led to a surge in research with two critical objectives: enhancing detection accuracy \cite{parmar_image_2018, liu2021swin} and minimizing model latency \cite{YOLO, yolox}. 
	Apparently, the simultaneous consideration of both objectives is vital for applications such as self-driving cars, where instant detection is necessary for immediate decision-making.
	Real-time object detection, also known as streaming perception, has thus gained attraction in recent years due to its role in comprehending the dynamic motion of the surrounding environment, and targets to do accurate detections with minimal delays, meeting the demands of real-time processing.
	
\begin{figure*}[t!]
		\centering
		\includesvg[width=0.95\textwidth]{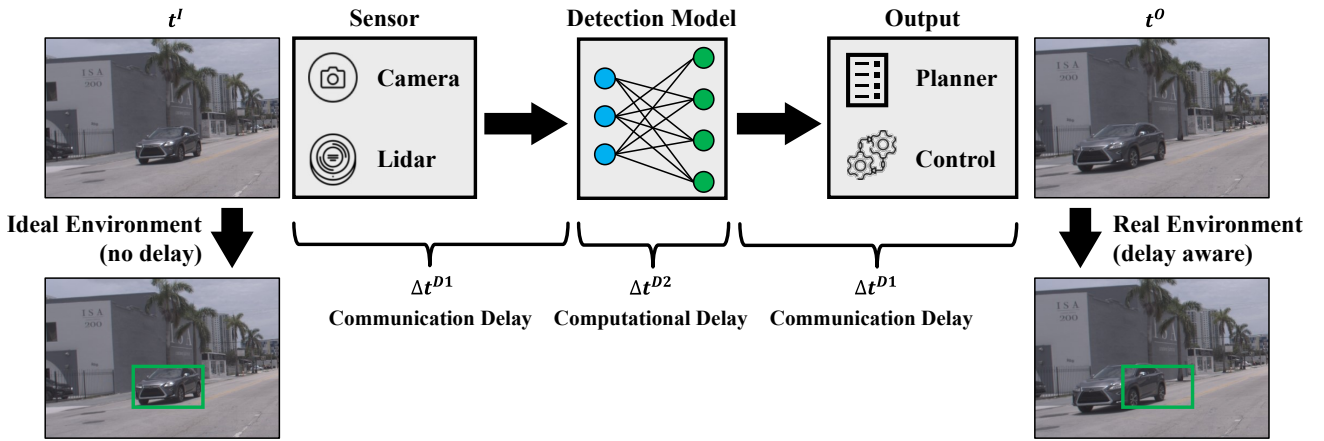}
		\caption{Demonstration of object displacement error in real world systems with potential communication-computational delays.}
		\label{fig:demo}
	\end{figure*}

	To meet real-time requirements, initial approaches attempted to enhance the speed of non-real-time detectors to achieve a higher frame rate of detection. However, as \cite{Towards_Streaming_Perception} highlighted, even with accelerated detectors, there are inherent delays in real-world applications.
	By the time a detector processes a frame, the environment has changed. This results in the temporal gap between the input of a frame and the output of a prediction, which can be fatal in scenarios where every millisecond counts, like autonomous driving.
	As shown in Figure \ref{fig:demo}, the detection is accurate in an ideal environment, where the detection process finishes instantly without any delays. However, a real-world environment would incur the non-negligible communication-computational delay in the detection process, causing a displacement between detection results (based on the input scenario) and the actual objects.
	
	
	
	In response to this critical issue, recent studies \cite{Towards_Streaming_Perception, StreamYOLO, DaDe, Longshortnet, DAMO-StreamNet} have augmented detectors with predictive capabilities, by training models to anticipate future objects' positions. \emph{However, these methods typically forecast object locations at a fixed interval (typically a single frame ahead), and expect the data communication together with the computation process to complete within the single frame interval \cite{StreamYOLO}}. For instance, a video stream with a frame rate of 30 would require the detection operation to complete within 1/30 of a second. So detectors must process the frame within strict time constraints, which means that most competent models strictly require low communication overhead and fast GPU processing \cite{StreamYOLO, Longshortnet, DAMO-StreamNet} even under diverse situations. 

	In this work, we observed and focused on an often-overlooked but significant issue: {\bf in the real world, the communication-computational delays vary significantly over time as communication bandwidth and workloads fluctuate and thus cannot be guaranteed to complete the whole detection process within a single frame.} As illustrated in Figure \ref{fig:delays}.(a), the same detector (the SOTA method DAMO-StreamNet \cite{DAMO-StreamNet}) faces widely different delays in one-frame inference across different devices. And even applied on the same device, the same detector experiences delays ranging from 27ms to 58ms depending on the system's workload as in Figure \ref{fig:delays}.(b). Delay fluctuations are also observed under different bandwidth limitations as shown in Figure \ref{fig:delays}.(c). This variability suggests that models may struggle to meet consistent real-time constraints during peak workloads or under low bandwidth, which could lead to critical failures in accurate real-time detection.
	
	
	
	
	To address this critical issue, this work introduces CorrDiff, a novel method that leverages temporal cues to integrate runtime information into the detection model. CorrDiff takes multiple past frames as input, producing predictions for multiple future frames simultaneously, while allowing for flexible output selection to adapt to delay fluctuations. The framework is specifically designed to tackle the challenges of real-time object detection under varying and dynamic delay conditions. The primary contributions of this work are summarized as follows.
	
	\begin{itemize}
		\item This work enables adaptive delay-aware streaming perception by integrating runtime information directly within detection models, concurrently predicting {multiple frames} and selectively produce the most appropriate output aligned with the real-world present time.
		\item Even in the realm of {single-frame} real-time detection, the proposed approach still outperforms all the existing state-of-the-art (SOTA) techniques.
		\item The proposed model demonstrates robust performance across a spectrum of devices, with GPUs that ranging from the high-performance Tesla V100 to the modest RTX 2080Ti. It achieves the highest level of perceptual accuracy on all platforms, whereas most SOTA methods struggle to complete computation within a single frame when applied on less powerful devices, thereby failing to meet the stringent real-time processing requirement.
	\end{itemize}
	
	For the rest of the paper, we will discuss relevant literature and potential gaps in Section \ref{sec:related_work}, the motivation of our approach in Section \ref{sec:motivation}, the methodology in Section \ref{sec:methodology} and the performance comparisons in Section \ref{sec:experiment}.
	
	\section{Related Work} \label{sec:related_work}
	
	{\bf Image Object Detection:}
	The evolution of deep learning has dramatically influenced the realm of object detection, with CNNs outpacing traditional methods. Image object detection methods are primarily divided into two-stage and one-stage methods. Two-stage methods are exemplified by R-CNN \cite{girshick_rich_2014}, which boosted accuracy by integrating region proposals with CNN features. Advancements in the two-stage approaches, seen in Fast R-CNN \cite{FastRCNN} and Faster R-CNN \cite{FasterRCNN}, streamlined the process by merging region proposal networks with CNN architecture. Nevertheless, these techniques still experience delays due to proposal refinement. In contrast, one-stage methods such as SSD \cite{SSD} and YOLO \cite{YOLO}, further removed the dependency on region proposals to reduce latency. They offered a balance of speed and precision for real-time tasks but lacked the temporal context necessary for streaming detection, as they focused solely on the current frame.
	
	{\bf Video Object Detection:}
	Video object detection (VOD) aims to detect objects on video data instead of static images. While early approaches processed each frame independently, they obviously failed to utilize video characteristics. Recent deep learning methods seek to make use of temporal-spatial relationship in the following means. Flow-based methods \cite{FGFA, MANet, THP, DFF} estimated optical flows to enhance the features of non-key frames. Tracking-based methods \cite{D&T} aimed to build object connections by learning feature similarities across frames. Attention-based methods \cite{RDN, SELSA, LLTR, HVRNet} applied attention mechanisms to establish temporal context relationships on long-duration videos. While these methods interpreted temporal information in various approaches, they generally focused on the offline setting, where detection delay is often overlooked. Contrastingly, this work takes both delay and temporal-spatial context into consideration, ensuring high accuracy with real-time performance.
	
	{\bf Streaming Perception:}
	Streaming perception aims to tackle the drift in real-time detection results caused by system latency. This drift is compensated by predicting entity locations after a certain number of frames using temporal information from historical results \cite{Towards_Streaming_Perception}. The sAP metric was introduced to evaluate object detection algorithms in streaming scenarios, factoring in both latency and accuracy. Chanakya \cite{Chanakya} attempted to improve sAP performance by learning a policy to alter input resolution and model size, efficiently balancing latency and accuracy. However, Chanakya did not base its method on a temporal-predictive detector. Subsequent researches thus introduced several models aimed at forecasting object locations. For example, StreamYOLO \cite{StreamYOLO} used a dual-flow perception module for next-frame prediction, combining both previous and current frames' features. Dade \cite{DaDe} and MTD \cite{MTD} introduced mechanisms to dynamically select features from past or future timestamps, taking the runtime delay of the algorithm into consideration. DAMO-StreamNet \cite{DAMO-StreamNet} and Longshortnet \cite{Longshortnet} used dual-path architectures to capture long-term motion and calibrate with short-term semantics. They extended dynamic flow path from 1 past frame to 3 past frames, successfully capturing the long-term motion of moving objects, and thus achieved the state-of-the-art performance in streaming perception. 
	{\bf These existing studies, however, only forecast objects' locations in the next frame, expecting the detector to finish computations within one frame.} Such approaches are infeasible on real-world devices, where potential communication-computational delays vary significantly and the whole detection process cannot be guaranteed within a single frame. Our proposed method is delay-aware and handles situations with high and dynamic lags. With the capability of using temporal cues inside detection model, our method is able to accommodates various communication-computational delays.
	
	\section{Motivation and Problem Formulation}
	\label{sec:motivation}
	
	Though communication-computational delays in real-time object detection have been studied in some research, the existing studies only produced a fixed-interval output (predicting the fixed one frame ahead \cite{StreamYOLO, DAMO-StreamNet}), and thus failed to adapt to the dynamic nature of delays. {In the following, we will show the observation of the real-world problems, which motivates this work.} 
	
	\begin{figure*}[t!]
		\centering
		\subfloat[]{\includegraphics[width=0.33\textwidth]{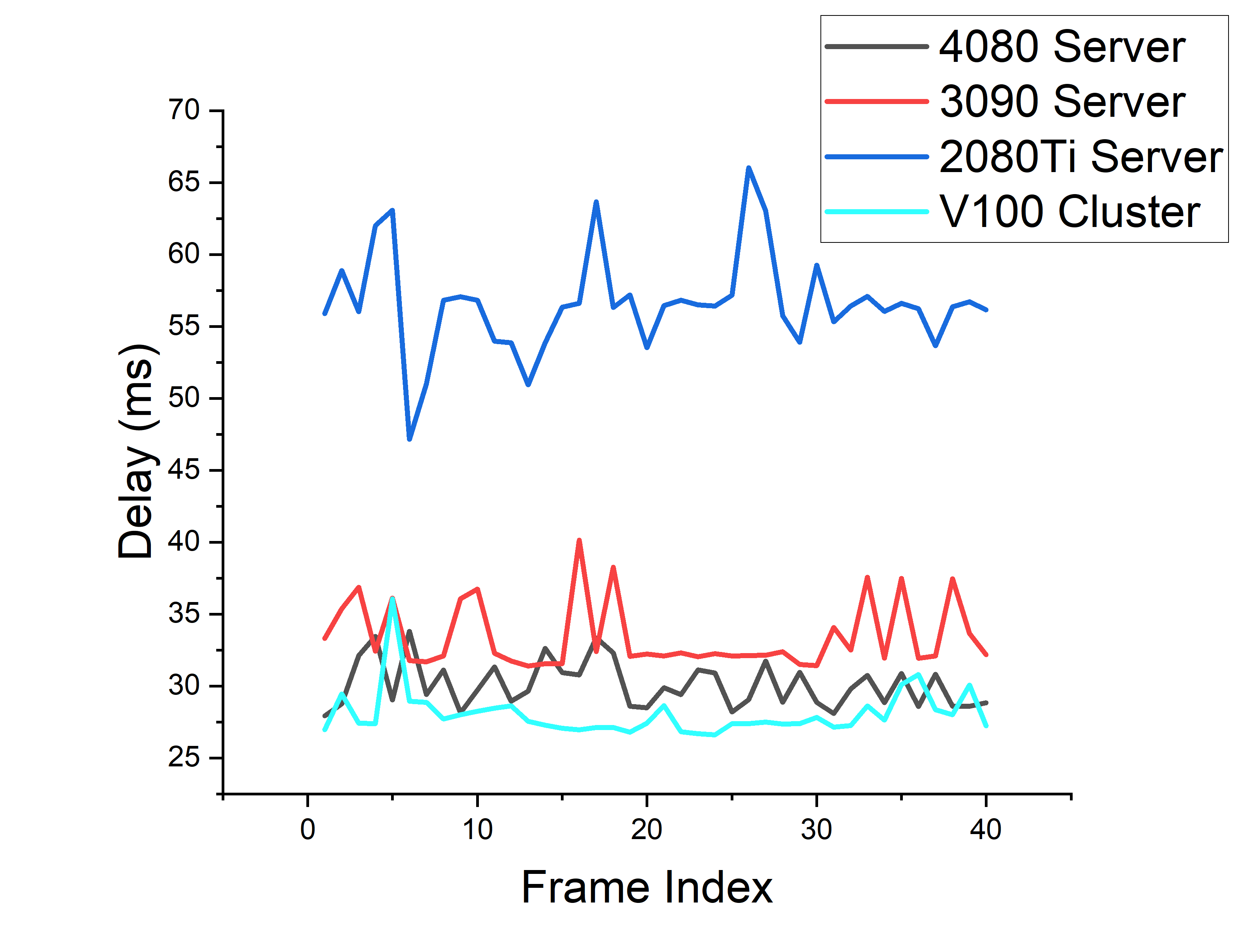}}
		\subfloat[]{\includegraphics[width=0.33\textwidth]{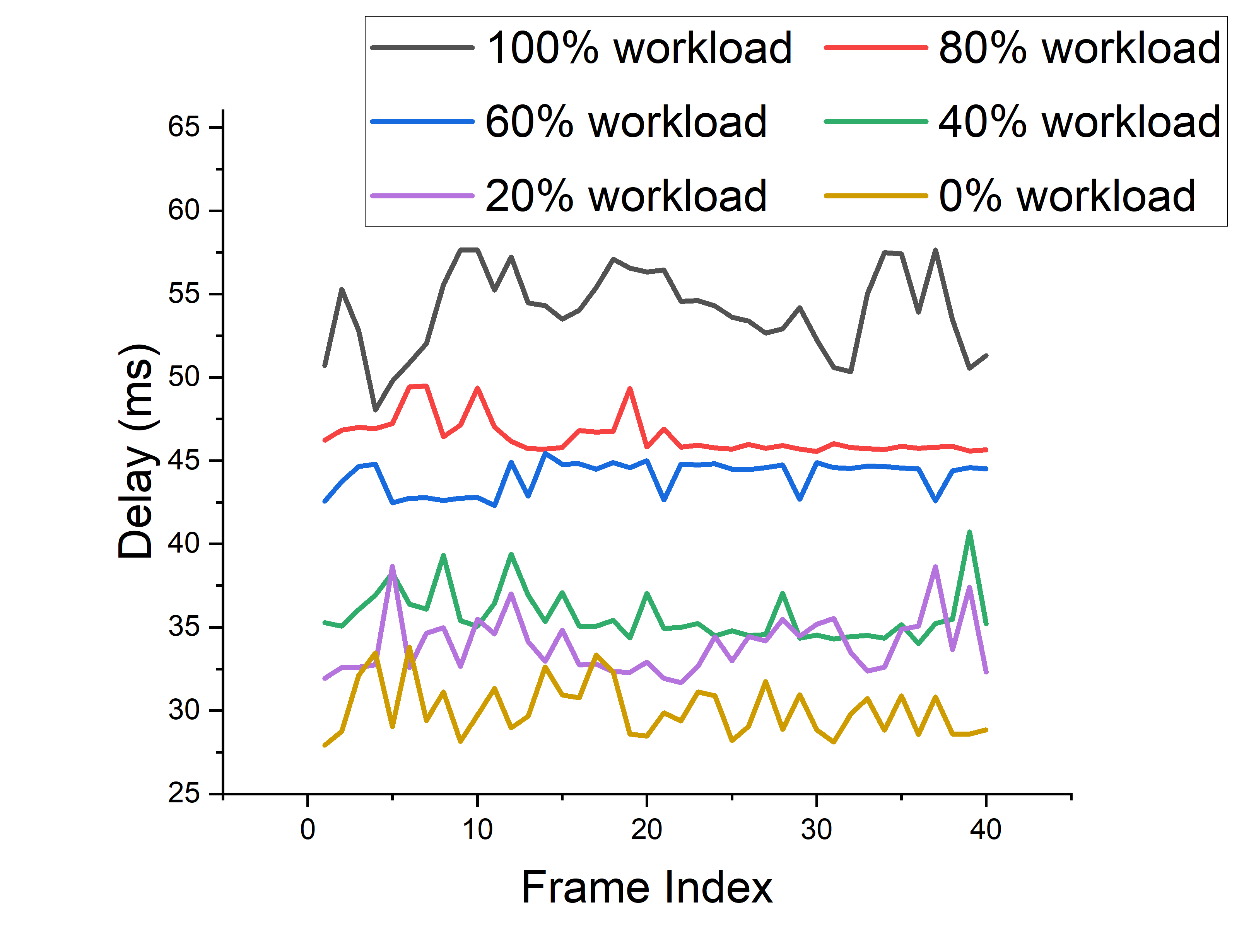}}
		\subfloat[]{\includegraphics[width=0.33\textwidth]{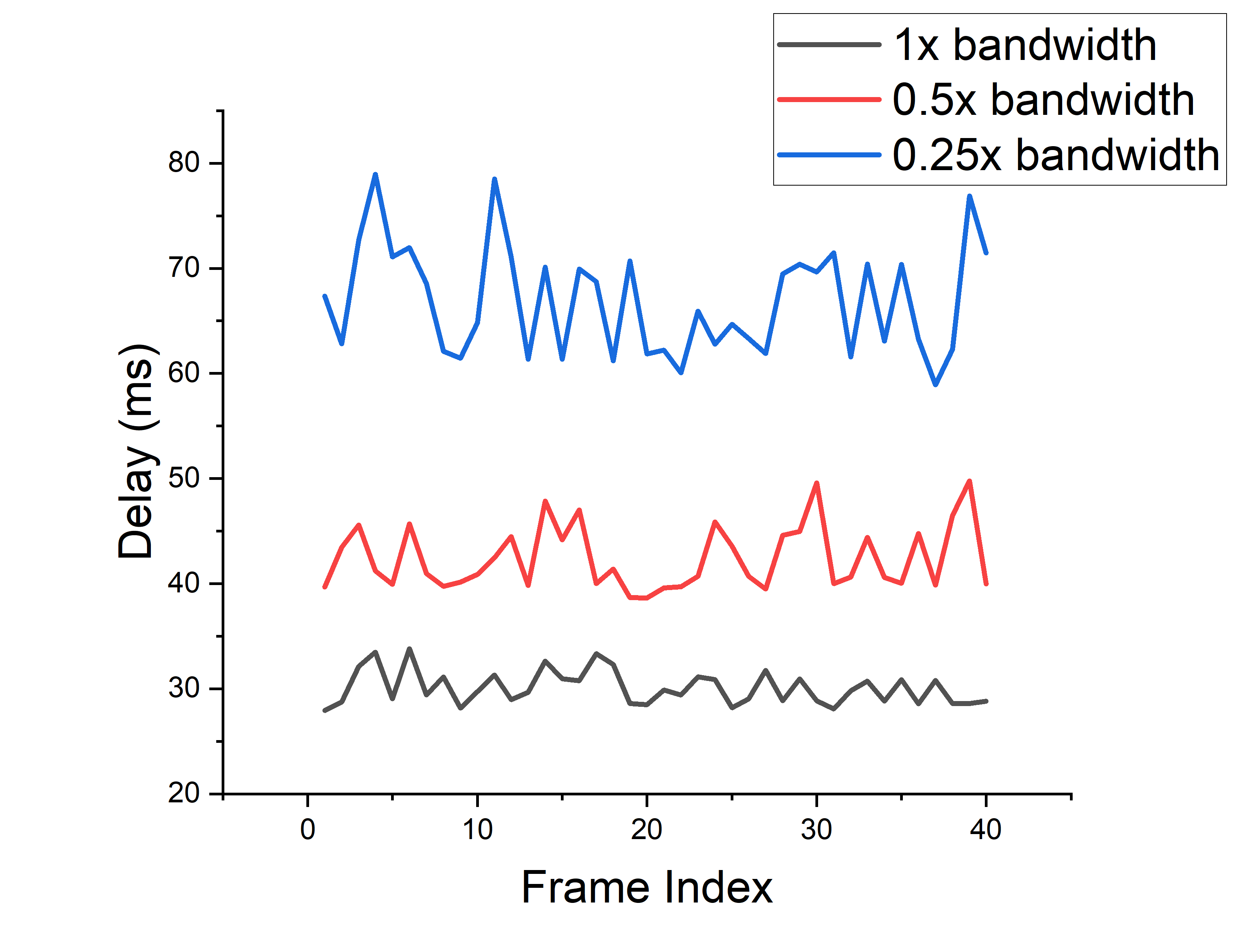}}
		\caption{One-frame inference delay of the DAMO-StreamNet \cite{DAMO-StreamNet} in different scenarios: (a) deploying on different devices. (b) deploying on a server with RTX 4080 but with various workloads (simulated by using a similar approach to GPU contention generation in \cite{Approxdet}). (c) deploying on a server with RTX 4080 but with different bandwidths.}
		\label{fig:delays}
	\end{figure*}
	
	\begin{figure*}[t!]
		\centering
		\includesvg[width=0.95\textwidth]{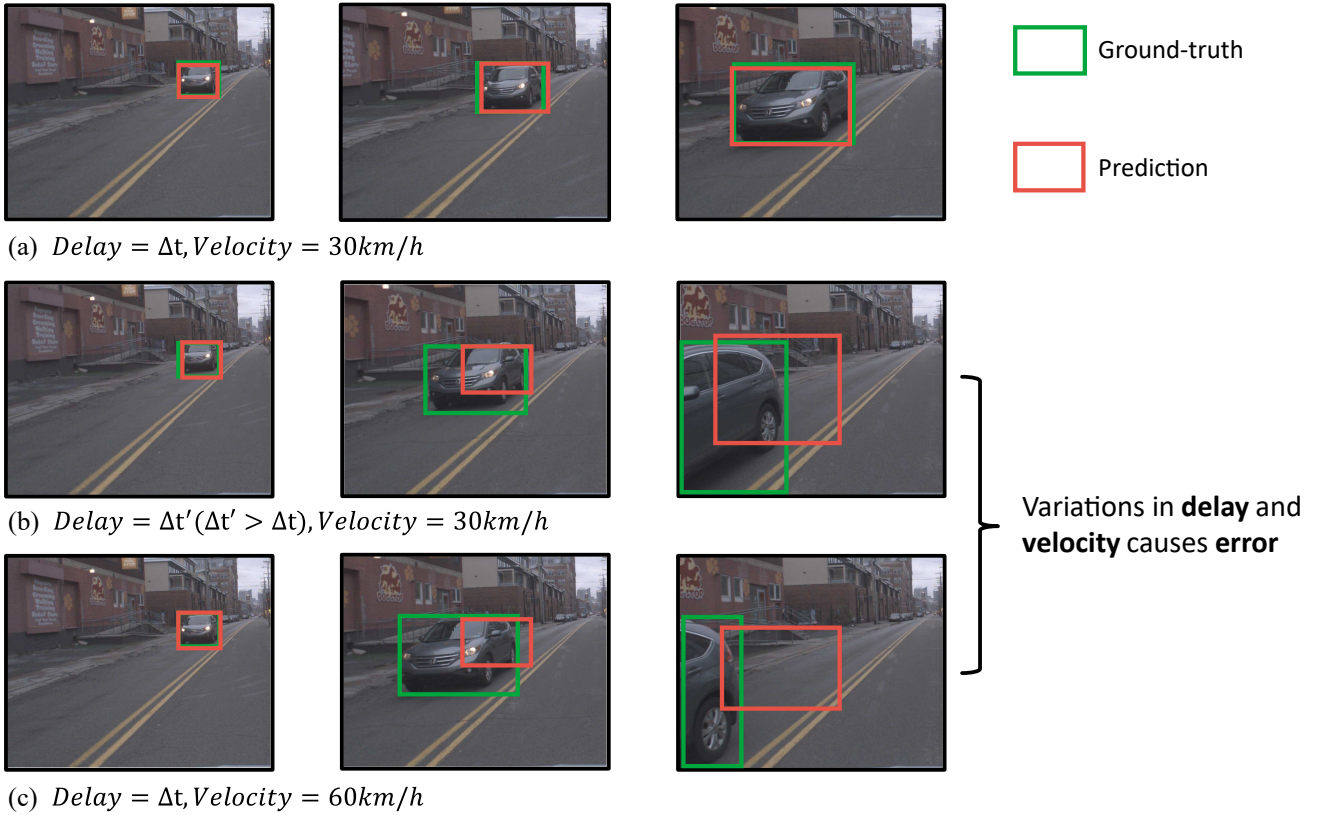}
		\caption{Motivation example: when a streaming detector is trained with fixed communication-computational delay (e.g. $\Delta t$) and fixed object velocities (e.g. $30km/h$), the inference can be accurate when (a) both delay and velocity remain the same as in training. But the detection accuracy will significantly decrease when either (b) the communication-computational delay varies (e.g. from $\Delta t$ to $\Delta t'$) or (c) the vehicle velocity changes (e.g. from $30km/h$ to $60km/h$).}
		\label{fig:motivation}
	\end{figure*}
	
	\subsection{Observation and Motivation}
	
	The existing studies in streaming perception \cite{StreamYOLO, Longshortnet, DAMO-StreamNet}, which focused on fixed frame prediction, 
	also have noticed that various object displacement may be involved by different moving velocity of vehicles, which can significantly affect the real-time detection. For instance, if the detector is only trained at a fixed frame rate (e.g., $60$ frames per second (FPS)), it can only be equipped to simulate vehicles moving at a corresponding constant speed (e.g., $30km/h$). As depicted in Figure \ref{fig:motivation}.(a) and Figure \ref{fig:motivation}.(c), when a vehicle's speed increases to $60km/s$, the detector (trained at the fixed frame rate to simulate $30km/h$) fails to accurately detect it.
	
	To address the above mentioned issue, training schemes with speed variation were applied in these work \cite{StreamYOLO, Longshortnet, DAMO-StreamNet}, and thus the trained model is capable to do detection on different speeds. However, they need to fix the speed before applying the model, i.e., at each frame $i$, determine that whether they do the detection for frame $i+1$ to simulate slow vehicle velocity or do detection for frame $i+3$ to simulate fast vehicle velocity. These approaches thus are inherently restrictive because they did not allow dynamic velocity of vehicles in runtime.
	
	In fact, we observed that dynamic and significant delays can be involved in one-frame detection for the detection model like DAMO-StreamNet \cite{DAMO-StreamNet}, because of various issues such as dynamic workloads or different bandwidth, as shown in Figure \ref{fig:delays}.(b) and Figure \ref{fig:delays}.(c). Other dynamic workloads could occupy GPU memory and utilization, affecting the computational delay of the model. Bandwidth variations could limit the speed of data and increase the communication delay.
	
	As a result, even the vehicle velocity remains the same (30km/h), as shown in Figure \ref{fig:motivation}.(a) and \ref{fig:motivation}.(b), different communication-computational delay will lead to inaccurate detection as it takes different duration to generate the output.
	It can be noted that various velocities and dynamic delays introduce a similar element of randomness during the detection process, complicating the prediction task.
	
	To address these challenges, we introduce CorrDiff, a novel approach that captures the inherent randomness in both vehicle velocity and communication-computational delays. Our solution is designed to provide real-time detection via dynamic prediction generation for multiple future frames utilizing temporal cues, adapting to various delays and object displacements. Specifically, CorrDiff is tailored to accommodate diverse runtime scenarios, thereby enhancing the safety and reliability of streaming perception systems in various operational contexts. 
	
	\subsection{Problem Formation}
	Given a real-time monocular video sequence $\{\mathrm{I}_i\} \in \mathbb{R}^{length \times 3 \times h \times w}$, the proposed method aims to generate the bounding box predictions $\{\hat{\mathrm{O}}_i\} \in \mathbb{R}^{length \times Objects \times 5}$ for each timestamp. During the streaming perception evaluation process, each frame is emitted at timestamp $t_i^{I}=\frac{i}{k}$, where $k$ is the frame rate (30 frames per second (FPS), typically). As described in Figure \ref{fig:demo}, while the detector is inferring the frames, potential delays will happen, which leads to a misaligned timestamp of the outputs $\{t_i^{O}\}$ compared to the inputs $\{t_i^{I}\}$. Generally, the delays compose of communication delay $\Delta t^{D1}$, computational delay $\Delta t^{D2}$ and optional start-up delay $\Delta t^{D3}$ (delay caused by not finishing the process of the previous frames before the current frame). Denote the sum of these delays as $\Delta t$, and thus we have
	
	\begin{equation}
		t_i^{O}=t_i^{I}+\Delta t_i=t_i^{I}+\Delta t_i^{D1}+\Delta t_i^{D2}+\Delta t_i^{D3}.
	\end{equation}
	
	The existing studies either assume $\Delta t_i = 0$ (the non-real-time detectors such as Fast-RCNN \cite{FastRCNN}), or assume it costs fixed number of frames, e.g. 1 frame (\cite{StreamYOLO, Longshortnet, DAMO-StreamNet}). For the latter case, they did prediction at $t_{i-1}^{I}$ to obtain the prediction of current frame $\hat{\mathrm{O}}_i$ at $t_{i-1}^{O}$, forcing the detection process to complete within one frame ($t_{i-1}^{O} < t_{i}^{I}$).
	
	An interesting and important observation from realistic system is that the delay is non-negligible and varies significantly. 
	This requires the algorithm to have the ability to forecast the objects' locations across multiple future frames, as various delays may occur. It should manage to compensate the object displacement error caused by communication delay $\Delta t^{D1}$, computational delay $\Delta t^{D2}$, and potential start-up delay $\Delta t^{D3}$.
	The algorithm's capabilities are formally defined as follows:
	\begin{itemize}
		\item For longer delays spanning more than a single frame, the detector needs to be capable of accurately predicting farther future across multiple frames. Denote the current frame as $\mathrm{I}_{i}$, given the past frames $\mathrm{I}_{i-1}, \mathrm{I}_{i-2}, \dots$, we generate a series of prediction $\hat{\mathrm{O}}_{i+1}, \hat{\mathrm{O}}_{i+2}, \dots$.
		\item With dynamic delays, the detector should be able to adaptively output the correct detection $\hat{\mathrm{O}}_{j}$ out of all the predictions, that aligned to the real world timing $t_{i}^{O}$. $\hat{\mathrm{O}}_{j}$ will be evaluated by comparing to ground-truth $\mathrm{O}_{j}$.
		\item Moreover, under realistic conditions, communication errors may render a frame $\mathrm{I}_{i}$ unavailable. The detector should still be able to generate predictions under such circumstance. 
	\end{itemize}
	
	Building on these requirements, we propose CorrDiff, a novel method that not only meets the aforementioned capabilities but also satisfies stringent real-time processing demands across various devices in realistic scenarios.
	
	\begin{figure*}[t!]
		\centering
		\includesvg[width=0.95\textwidth]{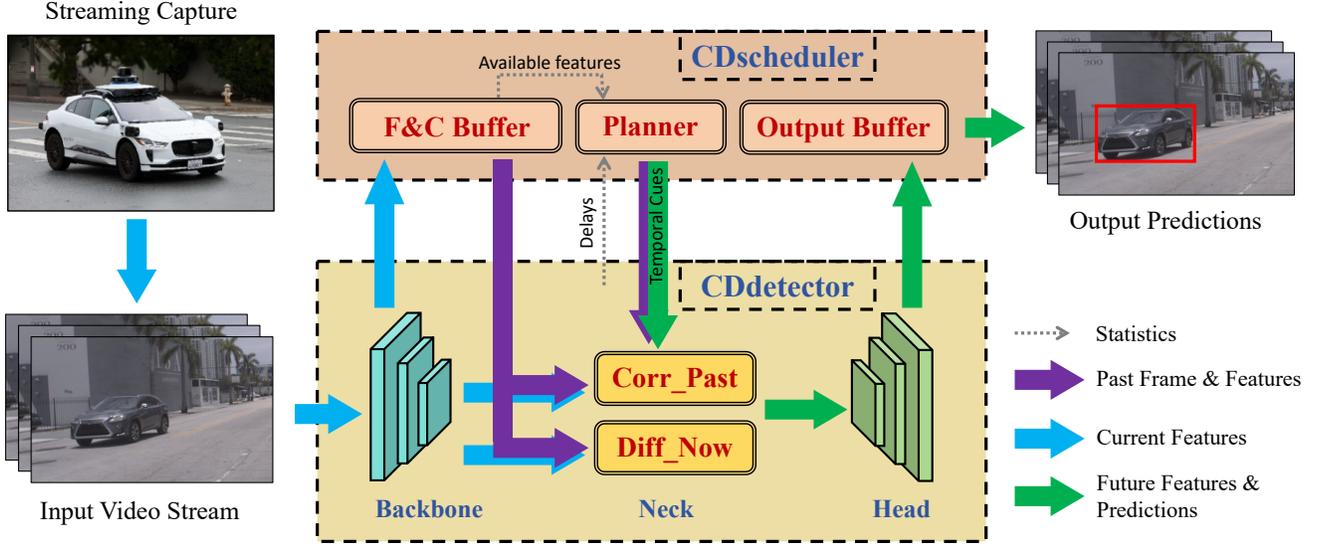}
		\caption{Overall architecture of CorrDiff. CorrDiff composes of a detection model CDdetector and a scheduling algorithm CDscheduler. CDdetector utilizes the Corr\_Past module and the Diff\_Now module, combining past and current features to produce future predictions. CDscheduler provides support by gathering runtime statistics to generate Temporal Cues, which is proceeded by CDdetector, making it adaptively delay-aware. The scheduler also uses 3 buffers: Historical Feature Buffer to reuse previously computed frame features, Corr\_Past Buffer to reuse correlation results and Output Buffer to store the freshest predictions and dispatch detection results at the corresponding timestamp. F\&C Buffer is the abbreviation for Historical Feature Buffer and Corr\_Past Buffer.}
		\label{fig:architecture}
	\end{figure*}

	\section{Methodology}
	\label{sec:methodology}
	
	In order to fill the research gap presented in current literature, we proposed CorrDiff, a detection system composes of a detection model named CDdetector and a scheduling algorithm named CDscheduler, to accommodate runtime delays and incorporate temporal cues inside the model. The detailed architecture is clarified in the following paragraphs.
	
	\subsection{Overall Design of CorrDiff}
	The general architecture of the proposed method CorrDiff is shown in Figure \ref{fig:architecture}. As a solution for real-time streaming detection, CorrDiff consists of 2 main modules, a CNN-based detection model CDdetector and a scheduling algorithm CDscheduler. CDdetector is capable of utilizing runtime temporal information, processing multiple past frames and generating predictions for multiple future frames. Meanwhile, CDscheduler assists CDdetector to adapt to various runtime conditions by collecting runtime statistics and providing CDdetector with temporal cues to guide its execution.
	
	In the following, for the detection model CDdetector, we firstly introduce 2 submodules named Corr\_Past and Diff\_Now to capture the temporal movement of observed objects. These 2 blocks utilizes past and future temporal cues, to fuse past frames' features and to predict the features for future frames. The past and future temporal cues denoted as $\mathrm{C^P}$ and $\mathrm{C^F}$ respectively, which indicates the indices of past frames and desired future predictions. At inference time, the model is paired with the scheduling algorithm CDscheduler to generate $\mathrm{C^P}$ and $\mathrm{C^F}$, determining the choice of input frames $\{\mathrm{I}_{j}\}, j \in \mathrm{C^P}$ and desired output predictions $\{\hat{\mathrm{O}}_{j}\}, j \in \mathrm{C^F}$. Next, for CDscheduler, we propose a method to compute temporal cues $\mathrm{C^P}$ and $\mathrm{C^F}$ from collected or estimated runtime statistics $\Delta \hat{t}^{D1}$, $\Delta \hat{t}^{D2}$ and $\Delta t^{D3}$. Apart from using runtime information, CDscheduler also uses Historical Feature Buffer and Corr\_Diff Buffer (together abbreviated as F\&C Buffer in Figure \ref{fig:architecture}) to hold previously generated image features and correlation features in a streaming fashion, in order to avoid recomputation and reduce $\Delta t^{D2}$. While the model outputs detection results for multiple future frames, the output predictions are stored inside Output Buffer and dispatched at its corresponding timestamp. In this buffer, newer or fresher predictions can update elder ones if they have not been dispatched.
	
	\subsection{CDdetector: The Detection Model}
	In the proposed method, we separate the design of CDdetector into 3 submodules: a DRFPN backbone for feature extraction, a CD neck for temporal feature fusing and forecasting, and a TAL head for predicting and decoding bounding boxes. Our work focuses on developing a neck module that utilizes temporal information to fuse multiple past frames' features, guiding the prediction of future objects' locations. We use the same backbone (DRFPN) and head (TAL) as in \cite{DAMO-StreamNet, StreamYOLO}. 
	
\begin{figure*}[t!]
		\centering
		\includesvg[width=0.95\textwidth]{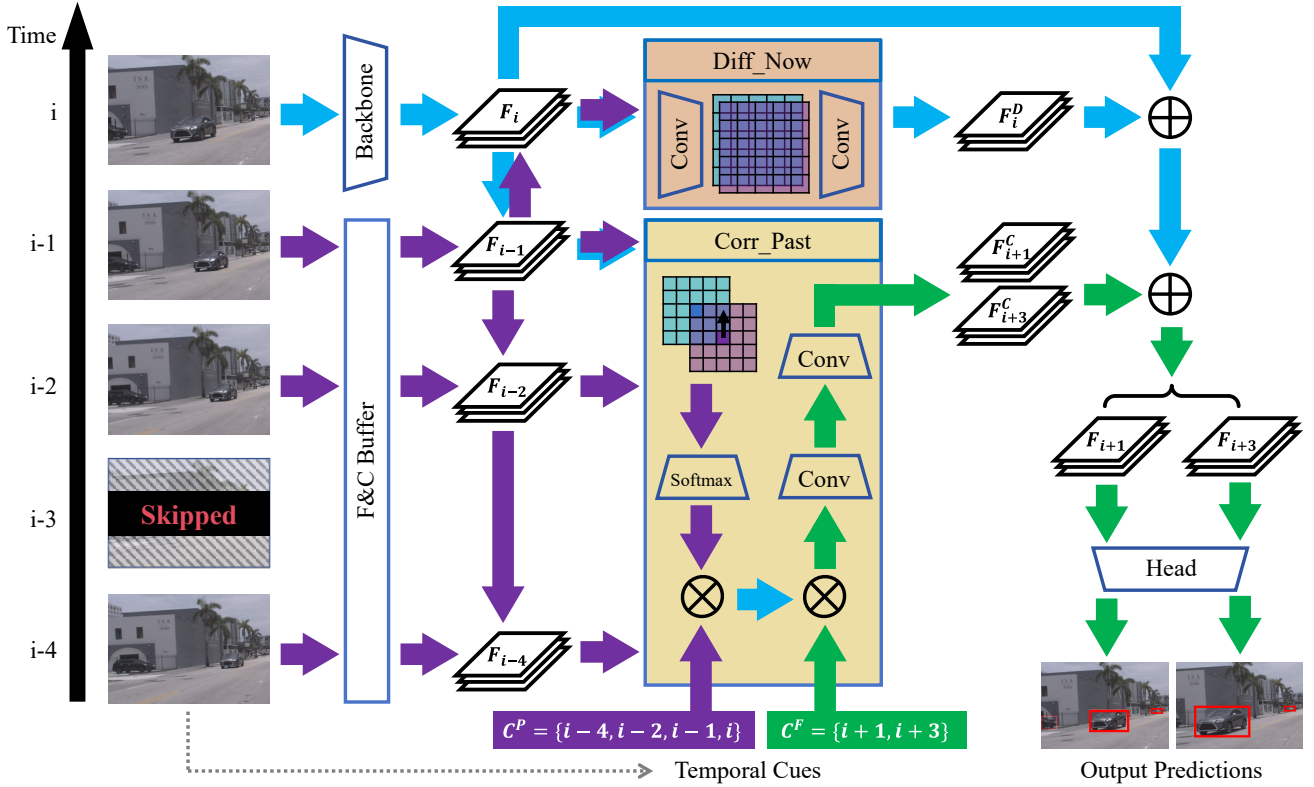}
		\caption{Detailed architecture of the detection model CDdetector. The example illustrates the situation that $i-3$ frame is skipped due to communication or computational delay. The detector is required to generate predictions for frame at $i+1$, $i+3$, given the image at $i$ and buffered features at $i-1$, $i-2$, $i-4$.}
		\label{fig:model}
	\end{figure*}

	At inference time, the model is given the current frame $\mathrm{I}_{i}$, buffered previous frames' features $\{\mathrm{F}_{j}\}, j \in \mathrm{C^P} \setminus \{i\}$ and the temporal cues $\mathrm{C^P}, \mathrm{C^F}$. CDdetector is required to generate future frames' predictions $\{\hat{\mathrm{O}}_{j}\}, j \in \mathrm{C^F}$. Firstly, the features of the current frame is extracted by the backbone module $\mathbf{W^B}(\cdot)$ and then concatenated to the buffered features $\mathcal{F}^P$ in a chronological order, producing past frames’ features $\mathcal{F}^P = \{\mathrm{F}_{j}\}, j \in \mathrm{C^P}$. Secondly, $\mathcal{F}^P$ is fed into the CD neck, which has 2 branches: Corr\_Past (Temporal Correlation, denoted as $\mathbf{W^C}(\cdot)$) and Diff\_Now (Temporal Difference, denoted as $\mathbf{W^D}(\cdot)$). Corr\_Past captures the potential movements of temporal features, while Diff\_Now enhances the quality of the current feature. These submodules merge past and current features to produce future frames' features $\mathcal{F}^F = \{\mathrm{F}_{j}\}, j \in \mathrm{C^F}$. Finally, the head module $\mathbf{W^H}(\cdot)$ process the future frames' features $\mathcal{F}^F$ in the same way as in ordinary object detection models, generating the estimated bounding boxes for future objects $\hat{\mathcal{O}}=\{\hat{\mathrm{O}}_{j}\}, j \in \mathrm{C^F}$. Note that when the current frame $\mathrm{I}_i$ is unavailable, detection process will start from the CD neck using previously stored buffers $\mathcal{F}^P$ and updated temporal cues $\mathrm{C^P}, \mathrm{C^F}$. The CDdetector can be described as
	
	\begin{subequations}
		\label{eq:model}
		\begin{align}
			\mathcal{F}^P &= \mathrm{Concat}(\mathbf{W^B}(\mathrm{I}_i), \{\mathrm{F}_{j}\}) \\
			\mathcal{F}^F &= \mathrm{Dup}(\mathrm{F}_i) + \mathbf{W^C}(\mathcal{F}^P,
			\mathrm{C^P}, \mathrm{C^F}) + \mathbf{W^D}(\mathcal{F}^P) \\
			\hat{\mathcal{O}} &= \mathbf{W^H}(\mathcal{F}^F),
		\end{align}
	\end{subequations}
	where $i$ denotes the index of current frame, $j \in \mathrm{C^F}$ denotes the index of past frames. $\mathrm{Concat}$ denotes tensor concatenation and $\mathrm{Dup}$ denotes feature duplication along the temporal dimension. The whole architecture of CDdetector is shown in Figure \ref{fig:model}.
	
	{\bf Corr\_Past:}
	The Corr\_Past module aims to capture the temporal dynamics of the past frame features. Inspired by optical flow models such as FastFlowNet \cite{FastFlowNet}, we utilize a spatial correlation sampler to measure the similarity between adjacent feature pixels across the temporal dimension. To ensure numerical stability, normalization is applied to prevent floating-point overflow. For each past feature $\mathrm{F}_{j}, j \in \mathrm{C^P} \setminus \{i\}$, we compute its correlation with with the feature of the next available past frame, $\mathrm{F}_{j'}$. For each point $x$ in $\mathrm{F}_{j}(x)$, we calculate its similarity in a local neighborhood of $\mathrm{F}_{j}(x)$, defined as $x+r_1$, where $ r_1 \in \mathcal{R}_1 = [-R_1, R_1] \times [-R_1, R_1]$. The similarity of 2 points is then determined by computing the dot product of their feature vectors and summing over a local region $\mathcal{R}_2 =[-R_2, R_2] \times [-R_2, R_2]$. Mathematically, the correlation of two frame features can be expressed as
	
	\begin{equation}
		\begin{split}
			& \mathrm{F^{C1}_j}(x) \\
			& = \underset{{r_1 \in \mathcal{R}_1}}{\mathrm{Concat}}(\mathrm{Similarity}(\mathrm{F}_j(x), \mathrm{F}_{j'}(x+r_1))) \\
			& = \underset{{r_1 \in \mathcal{R}_1}}{\mathrm{Concat}}(\sum_{r_2 \in \mathcal{R}_2} \mathrm{F}_j(x+r_1) \cdot \mathrm{F}_{j'}(x+r_1+r_2)),
		\end{split}
	\end{equation}
	where $\mathrm{Concat}$ denotes tensor concatenation along $\mathcal{R}_1$ and $\mathrm{Similarity}$ denotes similarity between 2 points. The output of the correlation operation is denoted as $\{\mathrm{F^{C1}_j}\}, j \in \mathrm{C^P} \setminus \{i\}$.
	
	To get an indicator of the overall movement of the current frame, we use weighted sum to fuse every $\mathrm{F^{C1}_j}$. The weights are calculated by the softmax of $\mathrm{C^P}$, the past temporal cue. We also multiply it with the future temporal cue $\mathrm{C^F}$ to get the movement of every future frame. The resulting correlation $\{\mathrm{F^{C2}_j}\}, j \in \mathrm{C^F}$ is then concatenated with $\mathrm{F}_i$ along the channel dimension, followed by 2 convolution operations to cast it back to its original number of channels. In practice, only the deepest feature from the feature pyramid is used. The correlation is up-sampled along spatial dimensions to match other feature in the feature pyramid. This method can capture movement in deep features with rich semantical context, lowering computation costs at the same time. The resulting feature is regarded as the correlation feature and denoted as $\{\mathrm{F}^C_{j}\}, j\in \mathrm{C^F}$.
	
	{\bf Diff\_Now:}
	Beside used to capture movement, past features $\{\mathrm{F}_{j}\}, j \in \mathrm{C^P}$ can also be used to enhance the features of the current frame $\mathrm{F}_i$. Adjacent frame's features can be used to clarify certain ambiguous areas in current frame's features. Diff\_Now firstly choose the most neighboring feature $\mathrm{F}_{max(j)}$ and both features pass a convolution block. The current feature is then subtracted by the neighboring feature, in order to distinguish the difference between these features. The resulting feature is regarded as the difference feature and denoted as $\mathrm{F}^D_i$. The process of Diff\_Now is described as
	
	\begin{equation}
		\begin{split}
			\mathrm{F^{D}_i} = \mathrm{Conv}_2(\mathrm{Conv}_1(\mathrm{F}_i) - \mathrm{Conv}_1(\mathrm{F}_{max(j)})),
		\end{split}
	\end{equation}
	where $\mathrm{Conv}_1$ and $\mathrm{Conv}_2$ denote convolution blocks.
	
	{\bf Combination:}
	While the correlation and difference features are computed, a residual connection from the current frame's feature is still needed. We argue that the current frame's feature is a reasonable initial guess of future frames' features. To this end, we firstly add $\mathrm{F}_i$ with $\mathrm{F}^D_i$, then duplicate it among the temporal dimension. The duplicated features are then added with $\{\mathrm{F}^C_{j}\}$ to produce $\{\mathrm{F}_{j}\}, j \in \mathrm{C^F}$, as the input to the head module. Finally, the CD neck can be mathematically represented as
	
	\begin{equation}
		\begin{split}
			\{\mathrm{F_j}\} = \mathrm{Dup}(\mathrm{F}_i + \mathrm{F}^D_i) + \{\mathrm{F}^C_{j}\},
		\end{split}
	\end{equation}
	where $i$ denotes the index of current frame and $j \in \mathrm{C^F}$ denote future indices.
	
	\subsection{CDscheduler: The Scheduling Algorithm}
	CDscheduler is responsible for generating temporal cues and arrange model execution, consisting of a Planner and three buffers. The Planner is the core of the algorithm. It collects and estimates runtime statistics to produce the temporal cues. The Planner also skips model execution when there is little time before the arrival of the following frame. The three buffers are: Historical Feature Buffer, Corr\_Past Buffer and Output Buffer. The first two buffers accelerate model execution by avoiding recomputation, while the Output Buffer dispatches multi-frame detection results to align with the real world timing and updates elder predictions with newer ones.
	
	\begin{figure*}[t!]
		\centering
		\includesvg[width=0.95\textwidth]{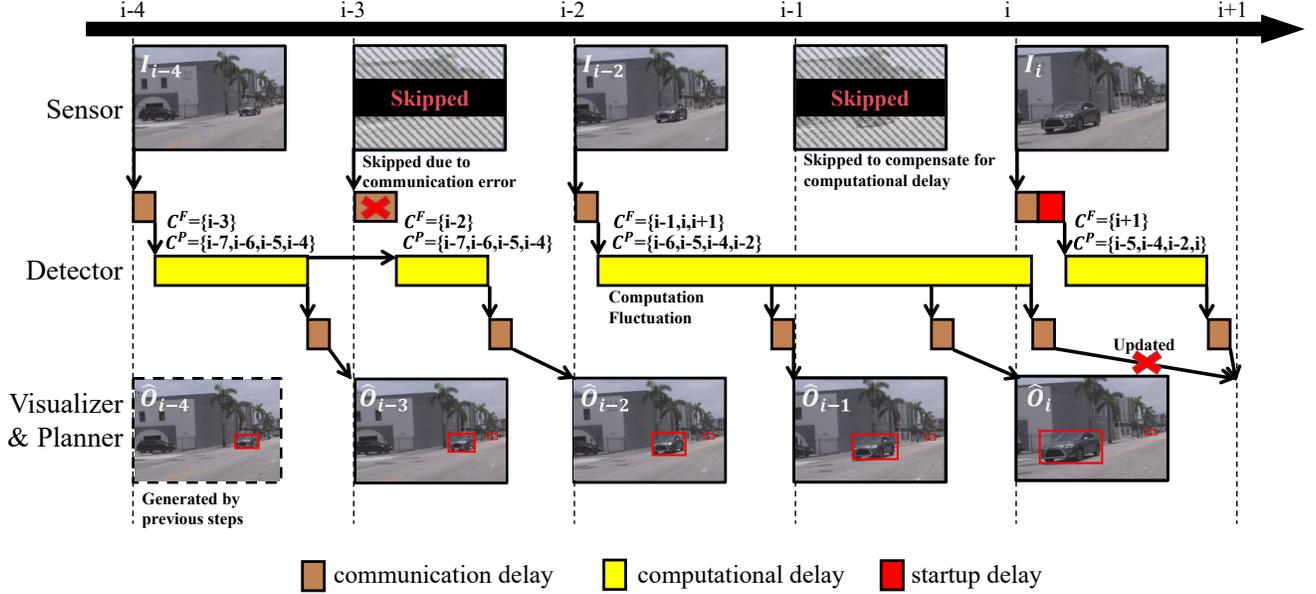}
		\caption{Demonstration of the scheduling algorithm CDscheduler. $\mathrm{C^P}, \mathrm{C^F}$ are temporal cues that determine the utilized past features and target future predictions of CDdetector. When communication errors or delays occur, the scheduler adjusts the temporal cues to let CDdetector skip unavailable frames and forecasts further into the future. It can be noted that the proposed method can guarantee accurate prediction output at each timestamp to satisfy real-time requirements.}
		\label{fig:schedule}
	\end{figure*}
	
	{\bf Planner:}
	The main function of the Planner is to generate an appropriate $\mathrm{C^P}$ and $\mathrm{C^F}$ so that the detection model's performance is maximized under Streaming Perception settings. During streaming inference, the communication delay, as well as the previous computation time of CDdetector's backbone, neck and head module, are firstly recorded as $\{\Delta t_{j}^{D1}, \Delta t_{j}^{D2B}, \Delta t_{j}^{D2N}, \Delta t_{j}^{D2H}\}, j < i$. As a reasonable guess, we estimate these delays in the current loop $\{\Delta \hat{t}_i^{D1}, \Delta \hat{t}_i^{D2B}, \Delta \hat{t}_i^{D2N}, \Delta \hat{t}_i^{D2H}\}$ with exponential moving average (EMA) of $\Delta t_{j}$ with a decay of 0.5. Additionally, the start-up delay of the current loop can be directly measured as $\Delta t_i^{D3}$. Therefore, the total estimated delay $\Delta \hat{t}_i$ can be denoted as:
	
	\begin{equation}
		\Delta \hat{t}_i =
		\begin{cases}
			\Delta \hat{t}_i^{D1} + \Delta \hat{t}_i^{D2B} + \Delta \hat{t}_i^{D2N} + \Delta \hat{t}_i^{D2H} + \Delta \hat{t}_i^{D3}, \\
			\hspace{1.5cm} \text{the current frame } \mathrm{I}_i \text{ is available} \\
			\\
			\sum_{j \in S}\Delta t_{j} + \Delta \hat{t}_i^{D1} + \Delta \hat{t}_i^{D2N} + \Delta \hat{t}_i^{D2H} + \Delta \hat{t}_i^{D3}, \\
			\hspace{1.5cm} \text{the current frame } \mathrm{I}_i \text{ is unavailable}
		\end{cases}
	\end{equation}
	where $S$ denotes the indices of other skipped adjacent frames. When the current frame is unavailable, we reuse previously computed past features so that the computational delay of the backbone $\Delta \hat{t}_i^{D2B}$ is not included in $\Delta \hat{t}_i$. To fully utilize past buffers, we select at most $m^P$ latest available buffers' indices as $\mathrm{C^P}$. Planner also limits the future predictions by producing at most $m^F$ predictions, indexed as $\mathrm{C^F}$. The future indices $\mathrm{C^F}$ approximate estimated output time $t_i^{I} + \Delta \hat{t}_i$. Finally, both $\mathrm{C^P}$ and $\mathrm{C^F}$ are clipped within the range of $[i-30, i+30]$ to keep temporal perception in reasonable range. The values are assigned empirically. This solution manages to maintain the balance between model latency and forecasting accuracy with respect to high and fluctuating delays.
	The demonstration of runtime statistics collection and temporal cue calculation are shown in Figure \ref{fig:schedule}. The figure depicts the situation where frame skips occurred due to communication error and computation delay. The Planner correctly produces the temporal cues to guide CDdetector into predicting frames that cover all timestamps.
	
	{\bf Historical Feature Buffer:}
	Though the features of past frames $\{\mathrm{F}_{j}\}, j \in \mathrm{C^P}$ can be directly acquired by backbone inference $\{\mathbf{W^B}(\mathrm{I}_{j})\}$, such method is not applicable in streaming evaluations, where the computation load clearly affects the final performance. Therefore, we cache historical frame's feature to avoid recomputation. We also set a limit to the length of the buffer and choose to discard the earliest buffered features after the buffer is full. Since early frames have a negligible influence on the current frame, the limitation ease the computation burden of the DRFPN backbone module, which has very large amount of parameters compared to other modules.
	
	{\bf Corr\_Past Buffer:}
	During the computation of Corr\_Past, we can observe that the correlation feature between previous frames can also be saved to avoid recomputation. Specifically, we can save the correlation features of the past frames $\{\mathrm{C}_{j}\}$. Since these features are irrelevant to the current feature, the following computations can reuse it.
	
	{\bf Output Buffer:}
	Although multiple future object predictions $\{\hat{\mathrm{O}}_{j}\}, j \in \mathrm{C^F}$ are produced by CorrDiff, they should not be emitted directly in sequential order.
	Because it may produce outdated result when $\Delta \hat{t}_i$ is underestimated. Therefore, we apply another buffer to the output of CorrDiff. At every timestamp, most temporally adjacent predictions in the buffer are dispatched, in order to effectively utilize all the predictions at appropriate timestamps. Furthermore, a newer prediction for index $j$ can update the older one if the older one still resides in the buffer. 
	
	\begin{figure*}[t!]
		\centering
		\includesvg[width=0.95\textwidth]{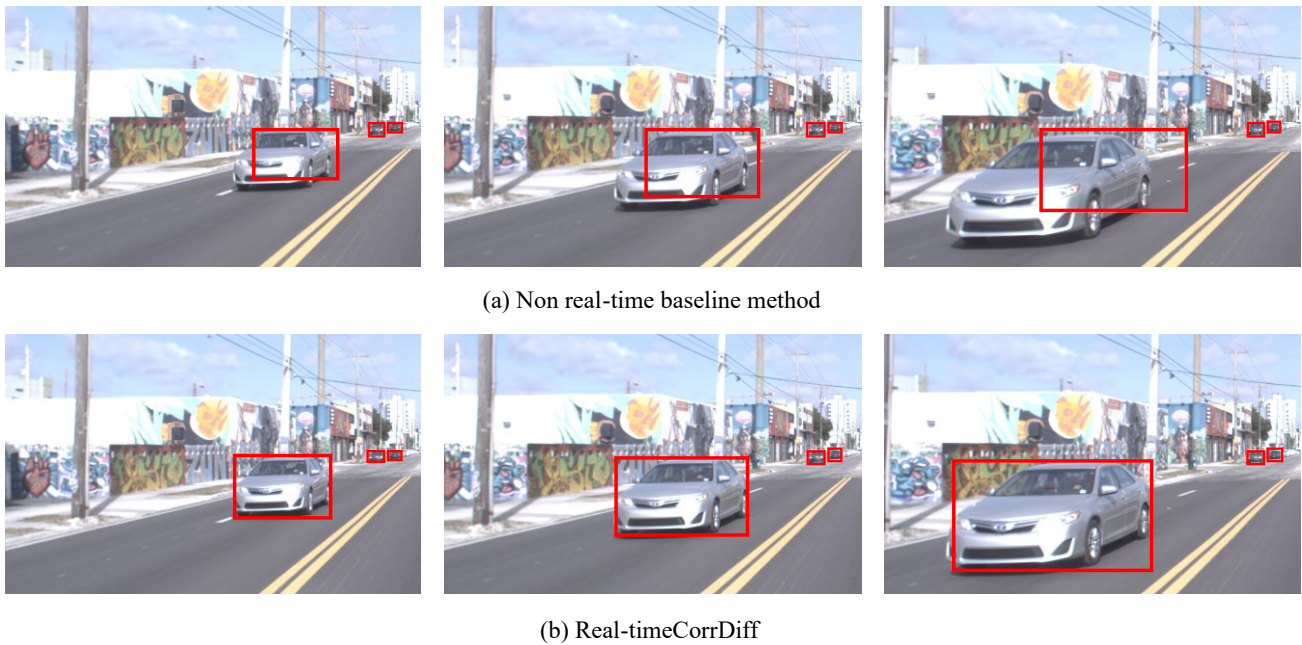}
		\caption{Demonstration of Non-real-time methods and CorrDiff on the Argoverse-HD dataset under streaming perception settings. Non-real-time methods (a) exhibits large object displacement error for its long communication and computation delay. CorrDiff (b) manages to produce up-to-date results for its low inference time and its ability to adapt to different delays and objects velocities.}
		\label{fig:result}
	\end{figure*}
	
	\subsection{Training and Inference}
	We mostly replicate the training scheme used by baseline methods \cite{DAMO-StreamNet, Longshortnet}, with the exception of Asymmetric Knowledge Distillation proposed by \cite{DAMO-StreamNet}, since we believe this contribution is orthogonal with our work.
	
	{\bf Mixed Speed Training:}
	To enhance model's delay-awareness, we also employ mixed speed training to extend the temporal perception range of CorrDiff. The model will be ineffective under other delays and velocities if only a fixed choice of temporal cues $\mathrm{C^P}$ and $\mathrm{C^F}$ is used for training. Therefore, we adopt mixed speed training scheme that samples $\mathrm{C^P}$ from $[-16, 0]$ and $\mathrm{C^F}$ from $[1, 16]$. The loss for each predicted future object is given equal weights.
	
	\section{Experiment} \label{sec:experiment}
	In this section, the implementations of both our method and evaluation metrics are elaborated. We also report the performances and ablation results of our method.
	
	\subsection{Implementation Details}
	{\bf Dataset:}
	We trained and tested CorrDiff on Argoverse-HD, an urban driving dataset composed of the front camera video sequence and bounding-box annotations for common road objects (e.g. cars, pedestrians, traffic lights). This dataset contains high-frequency annotation of 30 FPS that simulates real-world environment, which is suitable for streaming evaluation. We believe other datasets (e.g. nuScenes, Waymo) are not suitable for streaming evaluation as they are annotated at a lower frequency. We follow the train and validation split as in \cite{DAMO-StreamNet}.
	
	{\bf Model:}
	The base backbone of our proposed model is pretrained on the COCO dataset, which is consistent with the approach of \cite{DAMO-StreamNet}. Other parameters are initialized with Lecun weight initialization. The model is then fine-tuned on the Argoverse-HD dataset for 8 epochs using a single Nvidia GeForce RTX 4080 GPU with a batch size of 4 and half-resolution input ($600 \times 960$). To ensure a fair comparison with other methods, we provide 3 configurations of CorrDiff: CorrDiff-S (small), CorrDiff-M (medium) and CorrDiff-L (large). These configurations differ in the number of model parameters. The configuration setting is in accordance with previous methods' approaches \cite{DAMO-StreamNet, Longshortnet}.
	
	\begin{table*}
		\centering
		\tabcolsep=15pt\relax
		\caption{Main result of sAP comparison with real-time and non real-time SOTA detectors on the Argoverse-HD dataset. Best sAP scores are indicated in bold.}
		\label{tab: main}
		\begin{tabular}{lcccccc}
			\toprule
			Methods                           & $sAP$  & $sAP_{50}$ & $sAP_{75}$ & $sAP_S$ & $sAP_M$ & $sAP_L$ \\
			\midrule
			Non Real-time Methods & & & & & & \\
			\midrule
			Streamer (S=900) \cite{Towards_Streaming_Perception}                  & 18.2 & 35.3  & 16.8  & 4.7  & 14.4 & 34.6 \\
			Streamer (S=600) \cite{Towards_Streaming_Perception}                 & 20.4 & 35.6  & 20.8  & 3.6  & 18.0 & 47.2 \\
			Streamer + AdaS  \cite{Adascale, AdaptiveStreamer}                 & 13.8 & 23.4  & 14.2  & 0.2  & 9.0  & 39.9 \\
			Adaptive Streamer \cite{AdaptiveStreamer}                & 21.3 & 37.3  & 21.1  & 4.4  & 18.7 & 47.1 \\
			YOLOX-S  \cite{yolox}                         & 25.8 & 47.0  & 24.3  & 8.8  & 25.7 & 44.5 \\
			YOLOX-M  \cite{yolox}                         & 29.4 & 51.6  & 28.1  & 10.3 & 29.9 & 50.4 \\
			YOLOX-L  \cite{yolox}                         & 32.5 & 55.9  & 31.2  & 12.0 & 31.3 & 57.1 \\
			\midrule
			Real-time Methods & & & & & & \\
			\midrule
			StreamYOLO-S  \cite{StreamYOLO}                    & 28.8 & 50.3  & 27.6  & 9.7  & 30.7 & 53.1 \\
			StreamYOLO-M  \cite{StreamYOLO}                    & 32.9 & 54.0  & 32.5  & 12.4 & 34.8 & 58.1 \\
			StreamYOLO-L  \cite{StreamYOLO}                    & 36.1 & 57.6  & 35.6  & 13.8 & 37.1 & 63.3 \\
			DADE-L  \cite{DaDe}                          & 36.7 & 63.9  & 36.9  & 14.6 & 57.9 & 37.3 \\
			LongShortNet-S  \cite{Longshortnet}                  & 29.8 & 50.4  & 29.5  & 11.0 & 30.6 & 52.8 \\
			LongShortNet-M  \cite{Longshortnet}                  & 34.1 & 54.8  & 34.6  & 13.3 & 35.3 & 58.1 \\
			LongShortNet-L  \cite{Longshortnet}                  & 37.1 & 57.8  & 37.7  & 15.2 & 37.3 & 63.8 \\
			DAMO-StreamNet-S  \cite{DAMO-StreamNet}                & 31.8 & 52.3  & 31.0  & 11.4 & 32.9 & 58.7 \\
			DAMO-StreamNet-M  \cite{DAMO-StreamNet}                & 35.7 & 56.7  & 35.9  & 14.5 & 36.3 & 63.3 \\
			DAMO-StreamNet-L  \cite{DAMO-StreamNet}                & 37.8 & 59.1  & 38.6  & 16.1 & 39.0 & 64.6 \\
			CorrDiff-S(Ours)                  & 32.1 & 53.1  & 32.0  & 11.4 & 33.5 & 61.0 \\
			CorrDiff-M(Ours)                  & 36.0 & 57.5  & 36.2  & 14.6 & 36.9 & 64.6 \\
			CorrDiff-L(Ours)                  & \textbf{38.1} & \textbf{59.4}  & \textbf{39.0} & \textbf{16.6} & \textbf{39.5} & \textbf{65.7} \\
			\bottomrule
		\end{tabular}
	\end{table*}

	\begin{table} 
		\centering
		\tabcolsep=2pt\relax
		\caption{Delay Adaptation Metric comparison with different real-world devices. * means the model is trained using mixed speed training technique. Best sAP scores are indicated in bold.}
		\label{tab: compute}
		\begin{tabular}{llcccc}
			\toprule
			Devices                        & Methods        & $sAP^2$ & $sAP^4$ & $sAP^8$ & $sAP^{16}$ \\
			\midrule
			\multirow{3}{*}{v100 cluster}  & LongShortNet-S         & 25.5 & 21.6 & 16.7 & 11.0  \\
			& DAMO-StreamNet-S        & 25.0 & 19.9 & 14.2 & 9.4   \\
			& CorrDiff-S*(Ours) & \textbf{26.1} & \textbf{22.2} & \textbf{16.8} & \textbf{12.0}  \\
			\midrule
			\multirow{3}{*}{4080 server}   & LongShortNet-S         & 24.5 & 20.3 & 14.8 & 9.8   \\
			& DAMO-StreamNet-S        & 23.9 & 18.5 & 12.7 & 8.8   \\
			& CorrDiff-S*(Ours) & \textbf{25.0} & \textbf{20.8} & \textbf{15.6} & \textbf{11.0}  \\
			\midrule
			\multirow{3}{*}{3090 server}   & LongShortNet-S         & 24.0 & 20.0 & 14.4 & 9.5   \\
			& DAMO-StreamNet-S        & 23.7 & 18.2 & 12.6 & 8.8   \\
			& CorrDiff-S*(Ours) & \textbf{24.7} & \textbf{20.5} & \textbf{15.2} & \textbf{10.6}  \\
			\midrule
			\multirow{3}{*}{2080Ti server} & LongShortNet-S         & 23.0 & 18.0 & 12.7 & 8.6   \\
			& DAMO-StreamNet-S        & 21.9 & 16.4 & 11.3 & 8.0   \\
			& CorrDiff-S*(Ours) & \textbf{23.3} & \textbf{18.7} & \textbf{13.2} & \textbf{9.4}   \\
			\bottomrule
		\end{tabular}
		
	\end{table}
	
	{\bf Main Metric:}
	We follow the streaming evaluation methods proposed by \cite{Towards_Streaming_Perception} as the main test metric. The streaming Average Precision (sAP) is used to evaluate the performance of the whole pipeline under a simulated real-time situation. The sAP metric compares the output of the model with ground-truth at output timestamp, instead of input timestamp, which is common in offline evaluations. Specifically, the sAP metric realigns the prediction at time $t_i^{out}$ to index $j$ in order to match the indices of ground truth. Each index $j \in [0, L)$ is paired with the nearest prediction before the emission of frame $\mathrm{I}_j$. In other words, each ground truth item in $\{\mathrm{O}_j\}$ is paired with an item in detector output $\hat{\{\mathrm{O}_i\}}$, while satisfying
	
	\begin{equation}
		\label{eq:titj}
		t_i^{out} <= t_j^{in}, ~t_j^{in}<t_{i+1}^{out}.
	\end{equation}
	Note that it is possible that some consecutive ground truth frames are assigned to the same prediction frame, due to the fact that the delay costs more time than the frame time (33ms, typically). The prediction and ground-truth pairs are then evaluated via common detection metrics, such as mean Average Precision (mAP), which computes the average precision scores of matched objects with IoU thresholds ranging from 0.5 to 0.95. The sAP scores for small, medium, and large objects (denoted as $sAP_S, sAP_M, sAP_L$ respectively) are also reported. To ensure a fair comparison, we did not employ mixed speed training under the main metric. This is in accordance with the training scheme of baseline methods.
		
	\begin{table}
		\centering
		\tabcolsep=2pt\relax
		\caption{Acceleration Adaptation Metric comparison with different simulated speed variations. $2\times, 4\times, 8\times, 16\times$ of original vehicle speed are simulated by temporally downsampling frames, labeled as $mAP^2, mAP^4, mAP^8, mAP^{16}$, respectively. Best mAP scores are indicated in bold.}
		\label{tab: speed}
		\begin{tabular}{llcccc}
			\toprule
			Model sizes        & Methods                & $mAP^2$ & $mAP^4$ & $mAP^8$ & $mAP^{16}$ \\
			\midrule
			\multirow{4}{*}{S} & LongShortNet-S         & 26.4 & 20.6 & 13.4 & 8.9           \\
			& DAMO-StreamNet-S       & 28.9 & 22.0 & 14.6 & 9.4           \\
			& CorrDiff-S(Ours)       & \textbf{29.3} & 22.5 & 15.1 & 10.0           \\
			& CorrDiff-S*(Ours)      & 29.1 & \textbf{23.1} & \textbf{16.2} & \textbf{12.1}          \\
			\midrule
			\multirow{4}{*}{M} & LongShortNet-M         & 30.7 & 23.8 & 15.8 & 9.7           \\
			& DAMO-StreamNet-M       & 31.7 & 24.4 & 16.0 & 10.2          \\
			& CorrDiff-M(Ours)       & \textbf{32.1} & 24.9 & 16.6 & 11.1          \\
			& CorrDiff-M*(Ours)      & 31.9 & \textbf{25.4} & \textbf{17.2} & \textbf{12.8}          \\
			\midrule
			\multirow{4}{*}{L} & LongShortNet-L         & 33.2 & 25.8 & 17.2 & 10.7          \\
			& DAMO-StreamNet-L       & 33.8 & 26.1 & 17.1 & 10.8          \\
			& CorrDiff-L(Ours)       & \textbf{34.5} & 26.9 & 17.6 & 11.1          \\
			& CorrDiff-L*(Ours)      & 33.8 & \textbf{27.5} & \textbf{19.6} & \textbf{14.2}          \\
			\bottomrule
		\end{tabular}
	\end{table}
	
	{\bf Delay Adaptation Metric:}
	To simulate streaming perception on devices with diverse connectivity and computation capabilities, we test the framework on 1 online GPU computing cluster (denoted as v100 cluster) and 3 real-world servers (denoted as 4080 server, 3090 server and 2080Ti server) with different hardware specifications. The detailed information about these devices is listed in Table \ref{tab: hardware}. Additionally, we adopt a delay factor $d$ to simulate various delay situations, where all communication-computational delays are multiplied by $d$. We assign $d \in \{2, 4, 8, 16\}$ and denote corresponding sAP value as $sAP^d$.

	{\bf Acceleration Adaptation Metric:}
	To offer a more comprehensive comparison, we also evaluate the models (w/o strategy algorithm) under an offline setting using mean Average Precision (mAP). This approach ignores the impact of communication and computational delay, only simulates different amplitudes of vehicle acceleration by temporally downsampling frames. This metric tests the model's ability to handle various object displacements between adjacent frames. Unlike common offline evaluations, the prediction $\hat{\mathrm{O}}_i$ is evaluated against future objects $\{\mathrm{O}_{i+d}\}, d \in \{ 2, 4, 8, 16\}$. The resulting mAP scores are denoted as $mAP^2, mAP^4, mAP^8, mAP^{16}$, respectively.

	\begin{table*}[]
		\centering
		\caption{The hardware specifications of four devices used in the experiment.}
		\label{tab: hardware}
		\begin{tabular}{|l|l|l|l|}
			\hline
			\multicolumn{1}{|c|}{\textbf{Device Name}} & \multicolumn{1}{c|}{\textbf{CPU}} & \multicolumn{1}{c|}{\textbf{GPU}} & \multicolumn{1}{c|}{\textbf{Memory}} \\ \hline
			2080Ti server & Intel(R) Xeon(R) Gold 5118 CPU @ 2.30GHz × 48   & NVIDIA GeForce RTX 2080 Ti × 4 & 257547MB \\ \hline
			3090 server   & Intel(R) Core(TM) i9-10980XE CPU @ 3.00GHz × 36 & NVIDIA GeForce RTX 3090 × 2    & 128527MB \\ \hline
			4080 server   & Intel(R) Core(TM) i9-10900X CPU @ 3.70GHz × 20  & NVIDIA GeForce RTX 4080 × 2    & 257420MB \\ \hline
			v100 cluster  & Intel(R) Xeon(R) Gold 6226 CPU @ 2.70GHz × 24   & Tesla V100-SXM2-32GB × 8       & 256235MB \\ \hline
		\end{tabular}
	\end{table*}
	
	\begin{table}
		\centering
		\tabcolsep=8pt\relax
		\caption{Ablation study on design variations of CorrDiff-S. All the four components presented in this table are helpful in our framework. Since Planner and Buffer only function under streaming tests, we split the Ablation study into two parts evaluated by mAP and sAP, respectively. We also investigate different variations of the Corr\_Past and Diff\_Now modules. Note that the Buffer module is crucial in streaming perception settings by avoiding huge computation cost, both for CorrDiff and baseline methods. Best scores are indicated in bold.}
		\label{tab: ablation1}
		\centering
		\begin{tabular}{llccc}
			\toprule
			
			Corr\_Past & Diff\_Now & $mAP$  & $mAP_{50}$ & $mAP_{75}$ \\
			\midrule
			$\times$    & $\times$  & 29.0 & 50.1    & 29.1    \\
			$\times$    & concatenate & 30.8 & 51.9 & 30.6 \\
			$\times$    & addition & 31.2 & 52.3 & 31.1 \\
			$\times$    & \checkmark   & 31.5 & 52.6    & 31.5     \\
			channel-wise & \checkmark   & 31.8 & 52.9    & 31.7      \\
			\checkmark    & \checkmark   & \textbf{32.2} & \textbf{53.2}    & \textbf{32.1}  \\
			
			\midrule
			
			Planner & Buffer & $sAP$  & $sAP_{50}$ & $sAP_{75}$ \\
			\midrule
			$\times$       & $\times$      & 3.9  & 5.8     & 4.1 \\
			$\times$       & \checkmark      & 31.6 & 52.7    & 31.5 \\
			\checkmark       & \checkmark      & \textbf{32.1} & \textbf{53.2}    & \textbf{32.1} \\
			
			\bottomrule
		\end{tabular}
	\end{table}

	\begin{table}[]
		\centering
		\tabcolsep=8pt\relax
		\caption{Ablation study on different training speed of CorrDiff-S. $1\times, 2\times, 4\times, 8\times$ denotes training with a fixed speed rate $d \in \{1, 2, 4, 8\}$, respectively. $\mathrm{Mixed}$ denotes training with random speed rate $d$, sampling from $\{1, 2, 4, 8\}$ in every iteration. Best scores are indicated in bold.}
		\label{tab: ablation2}
		\begin{tabular}{lcccc}
			\hline
			Training Speed & $sAP^1$ & $sAP^2$ & $sAP^4$ & $sAP^8$ \\ \hline
			1x                      & \textbf{32.1}          & 30.1          & 24.7          & 17.6          \\
			2x                      & 28.1          & 26.1          & 22.5          & 15.7          \\
			4x                      & 18.4          & 16.5          & 14.4          & 11.3           \\
			8x                      & 12.6          & 10.1          & 9.4           & 7.9           \\
			Mixed                   & 31.6          & \textbf{30.3}          & \textbf{25.0}          & \textbf{17.9}          \\ \hline
		\end{tabular}
	\end{table}
	
	\subsection{Quantitative Results} 
	
	{\bf Overall Performance Comparison:}
	As the main result, our framework is evaluated against SOTA methods to demonstrate its strengths. Our proposed method achieves 38.1\% in sAP, as shown in Table \ref{tab: main}, surpassing the current SOTA method by 0.3\%. Under S and M configurations, our pipeline also achieves the first place in most metrics, compared to \cite{StreamYOLO}, \cite{Longshortnet} and \cite{DAMO-StreamNet}. The results clearly demonstrate the power of CorrDiff. Note that $sAP_L$ of our models is high, indicating that our proposed CorrDiff has recognized the temporal movement of large objects.
	
	{\bf sAP Comparison for Delay Adaptation:}
	To demonstrate the robustness of CorrDiff, we employ the Delay Adaptation Metric and evaluate both our framework and baseline models across 4 devices with 4 different delay factor $d$. As illustrated in Table \ref{tab: compute}, our method surpasses current SOTA DAMO-StreamNet \cite{DAMO-StreamNet} by a maximum of 2.6\% sAP on numerous real-world devices, demonstrating its strength on computation-bound environments. Notably, in high-latency scenarios, LongShortNet surpasses DAMO-StreamNet despite using a lighter backbone, indicating that large models do not necessarily scale effectively with increased latency. The result indicates that our method maintains its performance superiority even on devices with poor connectivity or low computing power.
	
	{\bf mAP Comparison for Acceleration Adaptation:}
	We also compared our method in simulated situation with drastic object accelerations. We evaluate our framework both with and without mixed speed training, to investigate the displacement adaptivity of the detection model. As shown in Table \ref{tab: speed}, our proposed method achieves an absolute improvement of 0.4\%, 1.1\%, 1.6\%, 2.7\% under $mAP^2$, $mAP^4$, $mAP^8$ and $mAP^{16}$, respectively, without mixed speed training compared to DAMO-StreamNet-S. Interestingly, although mixed-speed-trained models only have similar performance to other baseline models under $mAP^2$, a maximum of 3.4\% mAP gain is observed on larger temporal intervals. The experimental results demonstrate the temporal flexibility of our framework.
	
	{\bf Device Hardware Specifications}
	As shown in Table \ref{tab: hardware}, our experiment uses four different devices with variations in their communicational and computational capabilities. The Delay Adaptation Metric experiment is done on all four devices, while other experiments are done on the 4080 server. One thing to mention is that although we used different devices from those used in StreamYOLO \cite{StreamYOLO} and DAMO-StreamNet \cite{DAMO-StreamNet} (RTX 4080 vs Tesla V100), it does not affect the validity of our experiments. Generally, the V100 has similar or even slightly higher deep learning capacity (112 TFlops for V100 vs. 97.42 TFlops for 4080), so we believe the comparison to the SOTAs does not overestimate our method. Additionally, we also evaluated our S model on the V100, which achieved an sAP of 32.1, the same as our 4080 results, further validating the fairness of the comparison.

	\subsection{Ablation study}
	
	{\bf Design Variations:}
	The results of the ablation study are listed in Table \ref{tab: ablation1}. We verify the effectiveness of four proposed components: Corr\_Past, Diff\_Now, Planner and Buffers (Historical Feature Buffer, Corr\_Diff Buffer and Output Buffer). For Corr\_Past and Diff\_Now, we remove them from our model and observe a 0.7\% and 2.5\% decrease in test sAP, respectively. We also test other variations of Corr\_Past and Diff\_Now. For Corr\_Past, we tested other correlations such as per-channel correlation. For Diff\_Now, we tested concatenation or addition operation between the features. However, all substitutions are inferior to our current approach. As for Planner and Buffers which only operate under streaming settings, we examine their ability under $sAP$. We remove them from our framework and observe a 0.5\% and an astonishing 27.7\% decrease in the metric. The huge decrease in $sAP$ indicates that it is unrealistic to recompute past frames' feature under streaming perception settings.
	
	{\bf Mixed speed Training:}
	We also employ training under different velocities (by temporally downsampling frames) as well as mixed speed training (by sampling input frames from random frame intervals). The results in Figure \ref{tab: ablation2} show that fixing training speed will not necessarily increase sAP under simulated high-delay environments. Instead, mixed speed training increases accuracy in large $d$ values by randomly sampling different input frames' indices ($C^P$), potentially modeling the objects' movement under different velocities.
	
	\newpage
	
	\section{Discussion}
	
	{\bf Conclusion:}
	We introduce CorrDiff, a novel streaming perception framework that utilizes temporal cues to produce multiple predictions that aligns with real-world time, effectively producing real-time detection results. CorrDiff is the pioneer framework in streaming perception that make use of temporal cues and multi-frame output. Inspired from optical flow models, we design the detection model that handles input and output with a dynamic temporal range. The scheduling algorithm is also proposed to provide buffer techniques and temporal cues to the detector. Our method not only outperforms current SOTA methods under ordinary streaming perception settings, but also surpasses these methods by a large margin under high/dynamic communication-computational delay \& drastic object acceleration environments.
	
	{\bf Limitation:}
	Despite the demonstrated strengths, CorrDiff still has its limitations. First, though temporal cues $\mathrm{C^P}$ and $\mathrm{C^F}$ are produced to provide runtime information, the detection model only utilizes it in the computation of Corr as a coefficient. A stronger integration could be achieved if it further impacts the computation of features. Second, our Delay Adaptation Metric will sometimes not reflect the true ability of the model due to temporal aliasing effect. In other words, the model will perform poorly if computational time slightly misaligns with multiples of frame rate. For example, if the model's total delay time is slightly over one frame interval (say, 34 ms), at time $i$, its prediction will be evaluated with ground truth $\mathrm{O}_{i+2}$. If the delay time is 32 ms, its prediction will instead be compared with $\mathrm{O}_{i+1}$, making a large difference from a slight delay change. Therefore, we should sample more values of $d$ (decimal values, instead of integers) to concretely evaluate the model. Third, although we considered the fluctuation of the model's computation time, the evaluation of impacts from different external workloads is not included in our experiment. The reason is that the evaluation may not be deterministic, since the model and the workload could affect each other. E.g., the model could cause the workload's GPU operations to queue, reducing the workload's impact on the model. This interference could introduce uncertainty that makes the results unreproducible. We leave these limitations for future work.
	
	\newpage
	
	\bibliography{egbib}{}
	\bibliographystyle{plain}
	
\end{document}